\documentclass[acmlarge]{acmart}

\usepackage{color,array}

\usepackage{graphicx}

\usepackage{amsmath,amsfonts}
\usepackage{algorithmic}
\usepackage{algorithm}
\usepackage{array}
\usepackage[caption=false,font=normalsize,labelfont=sf,textfont=sf]{subfig}
\usepackage{textcomp}
\usepackage{stfloats}
\usepackage{verbatim}
\usepackage{graphicx}

\usepackage{longtable} 

\usepackage{multirow}
\usepackage{booktabs}

\usepackage{tikz}
\usepackage{array}

\usepackage[table]{xcolor} 
\usepackage{colortbl}      
\usepackage{makecell}      
\usepackage{pifont}        
\definecolor{modelcolor2}{HTML}{dad2e9} 
\definecolor{modelcolor3}{HTML}{fce5cd} 
\colorlet{rowblue}{gray!60!blue!10!white}

\newcommand{\checkmarkcolor}{%
    \tikz[baseline=(checkIcon.base)]{%
        \node[draw=gray!30, line width=0.5pt, fill=green!70!black, rounded corners, inner sep=1.5pt] (checkIcon) {\textcolor{white}{$\checkmark$}};%
    }%
}

\newcommand{\xmarkcolor}{%
    \tikz[baseline=-0.5ex, scale=0.2]{%
        \draw [line width=1.5pt, color=red] (0,0) -- (1,1);
        \draw [line width=1.5pt, color=red] (1,0) -- (0,1);
    }%
}

\newcolumntype{C}[1]{>{\centering\arraybackslash}m{#1}} 
\newcolumntype{L}[1]{>{\raggedright\arraybackslash}p{#1}} 

\AtBeginDocument{%
  }

\setcopyright{acmlicensed}
\copyrightyear{2026}
\acmYear{2026}
\acmDOI{XXXXXXX.XXXXXXX}

\acmJournal{POMACS}
\acmVolume{37}
\acmNumber{4}
\acmArticle{111}
\acmMonth{8}

\begin{document}

\title{Hallucination in Multimodal Foundation Models: A Survey on Causes, Corrections, and Evaluations}

\author{Yinghao Guo}
\thanks{This paper was supported by the National Natural Science Foundation of China (U24A20256); the Natural Science Foundation of Guangxi (2024GXNSFFA010006); the Guangxi Bagui Youth Talent Program; the Project of Guangxi Key Laboratory of Eye Health (GXYJK-202407); the Project of Guangxi Health Commission Eye and Related Diseases Artificial Intelligence Screen Technology Key Laboratory (GXYAI-202402).}
\affiliation{%
  \department{School of Computer, Electronic and Information}
  \institution{Guangxi University}
  \city{Nanning}
  \postcode{530004}
  \country{China}
}
\affiliation{%
  \institution{Guangxi Academy of Artificial Intelligence}
  \city{Nanning}
  \postcode{541004}
  \country{China}
}
\orcid{0000-0003-0027-8574}
\email{yinghao_guo@st.gxu.edu.cn}

\author{Wei Lan}
\affiliation{%
  \department{School of Computer, Electronic and Information}
  \institution{Guangxi Key Laboratory of Multimedia Communications and Network Technology, Guangxi University}
  \city{Nanning}
  \postcode{530004}
  \country{China}
}
\orcid{0000-0001-5839-7504}
\email{lanwei@gxu.edu.cn}

\author{Wenyi Chen}
\affiliation{%
  \department{School of Computer, Electronic and Information}
  \institution{Guangxi University}
  \city{Nanning}
  \postcode{530004}
  \country{China}
}
\orcid{0009-0007-3229-9341}
\email{wenychen@st.gxu.edu.cn}

\author{Qingfeng Chen}
\affiliation{%
  \department{School of Computer, Electronic and Information}
  \institution{Guangxi University}
  \city{Nanning}
  \postcode{530004}
  \country{China}
}
\orcid{0000-0002-5506-8913}
\email{qingfeng@gxu.edu.cn}

\author{Shichao Zhang}
\affiliation{%
  \institution{Guangxi Key Laboratory of Multi-Source Information Mining Security, Guangxi Normal University}
  \city{Guilin}
  \postcode{541004}
  \country{China}
}
\orcid{0000-0003-4052-1823}
\email{zhangsc@mailbox.gxnu.edu.cn}

\author{Shirui Pan}
\authornote{Corresponding author.}
\affiliation{%
  \department{School of Information and Communication Technology}
  \institution{Griffith University}
  \city{Southport}
  \state{QLD}
  \postcode{4215}
  \country{Australia}
}
\orcid{0000-0003-0794-527X}
\email{s.pan@griffith.edu.au}

\author{Huiyu Zhou}
\affiliation{%
  \department{School of Computing and Mathematical Sciences}
  \institution{University of Leicester}
  \city{Leicester}
  \postcode{LE1 7RH}
  \country{U.K.}
}
\orcid{0000-0003-1634-9840}
\email{hz143@leicester.ac.uk}

\author{Yi Pan}
\affiliation{%
  \department{School of Faculty of Computer Science and Control Engineering}
  \institution{Shenzhen University of Advanced Technology}
  \city{Shenzhen}
  \postcode{518107}
  \country{China}
}
\orcid{0000-0002-2766-3096}
\email{panyi@suat-sz.edu.cn}


\renewcommand{\shortauthors}{Guo et al.}

\begin{abstract}
Multimodal Foundation Models represent a significant leap in artificial intelligence. Among them, Large Vision-Language Models (LVLMs) serve as the typical representative of these foundation models, which integrate visual modality directly into Large Language Models (LLMs). They have demonstrated strong capabilities in information processing and generation. However, the existence of hallucinations has limited the potential and practical effectiveness of LVLM in various fields. Although lots of work has been devoted to hallucination mitigation and correction, there are few reviews to summarize them. To address this gap, this survey provides a systematic review of the hallucination landscape in LVLMs. We categorize the causes related to model architecture and data quality, and construct a comprehensive taxonomy of existing mitigation strategies. Furthermore, we critically assess current hallucination evaluation benchmarks from both discriminative and generative perspectives, highlighting the limitations of existing metrics. This survey concludes by discussing open challenges and future research directions to advance the reliability and trustworthiness of LVLMs.
\end{abstract}

\ccsdesc[500]{General and reference~Surveys and overviews}
\ccsdesc[500]{Computing methodologies~Natural language generation}
\ccsdesc[500]{Computing methodologies~Natural language processing}
\ccsdesc[500]{Computing methodologies~Computer vision problems}
\ccsdesc[300]{Computing methodologies~Neural networks}


\keywords{Large Visual Language Models, Hallucination Correction, Hallucination Evaluation Benchmarks.}


\maketitle

\section{Introduction}
{I}{n} recent years, LLMs have achieved excellent results in the field of natural language processing (NLP). Transformer-based LLMs acquire the ability to understand and generate natural language by learning the linguistic patterns and knowledge on a large-scale corpus. Lots of LLMs have emerged in the field of NLP such as GPT-4 \cite{15}, LLaMa \cite{62},  InstructGPT \cite{101},  PaLM \cite{102} and  Vicuna \cite{65}. Supported by the large-scale corpus and huge number of parameters, these LLMs can accomplish a wide range of tasks and have shown powerful zero-shot capabilities.

Although LLMs have exciting and robust properties, LLMs are limited to the text-only domains. Increasing works have been proposed to integrate visual information to LLMs. These new models are called LVLMs which  can be used in a variety of applications \cite{110, 111}, such as medical diagnosis and assistance \cite{80,81}, arts and entertainment \cite{82}, autonomous driving \cite{83}, virtual assistants and chatbots \cite{55,100}. With their surprising performance, LVLMs have attracted many users.
However, some users have found that LVLMs may generate information that is factually incorrect but appears plausible, such as incorrectly reporting nonexistent objects, object attributes, behaviors, and relationships between objects \cite{10.1145/3786319}.
The above phenomenon is known as hallucination which leads to the inability of LVLMs to be applied in scenarios with high accuracy and reliability \cite{10.1145/3777382}. For example, hallucinations may mislead users with incorrect or inaccurate information and even lead to the dissemination of misinformation in content summarization or information retrieval \cite{pan-etal-2023-risk}. If the LVLMs frequently generate hallucinations, it may affect the development of LVLMs \cite{10.1145/3712001}|. Therefore, correcting or mitigating hallucinations is necessary for LVLMs.

In order to build a trustworthy LVLM, the hallucination is a obstacle need to be overcame. As a result, a number of efforts have emerged to mitigate or correct the hallucinations of LVLM. Currently, several surveys have summarized the hallucination correction work in LLMs \cite{103,104}. In the realm of multi-modality, there has partial work\cite{105,106} aim to summary the hallucinatory phenomena of multimodal large language models. However, our survey employs a distinctly different taxonomic strategy. We categorize by the core ideas of various hallucination correction efforts and hallucination assessment benchmarks. 


\paragraph{\textbf{Our Contributions.}}
This paper presents a comprehensive review of the hallucination landscape in Large Vision-Language Models. To distinguish our work from existing surveys, we make the following key contributions:

\begin{itemize}
    \item \textbf{Systematic Analysis of Causes.} 
    We conduct a deep analysis of the origins of hallucinations in LVLMs. We categorize the causes of these issues into three main dimensions: multimodal fusion challenges, low-quality training data, and inherent model uncertainty. Furthermore, we explicitly analyze how modality gaps, data noise, and decoding randomness constitute the underlying causes of these failures.

    \item \textbf{Novel Taxonomy for Mitigation Strategies.} 
    We propose a structured taxonomy to categorize existing hallucination correction methods. We classify these strategies into three distinct phases: Learning from Low-quality Data, Addressing Fusion Challenges, and Post-Fusion Rectification. Within this framework, we systematically review representative techniques such as data rewriting, perceptual reinforcement, and reinforcement learning from human feedback.

    \item \textbf{Comprehensive Evaluation Benchmarks.} 
    We critically assess the current benchmarks for measuring hallucinations. We categorize evaluation paradigms into Discriminative Benchmarks and Generative Benchmarks. The former verifies fine-grained visual details and parametric knowledge, while the latter assesses free-form responses and robustness against adversarial inputs.

    \item \textbf{Insightful Future Directions.} 
    We identify critical gaps in current research and propose four promising directions. We advocate for establishing a unified theory for fusion uncertainty and shifting towards data-centric training. Additionally, we highlight the urgent need for automated evaluation pipelines and dynamically evolving correction frameworks.
\end{itemize}


\section{Background}

\begin{figure*}[!t]
\centering
\includegraphics[width=6in]{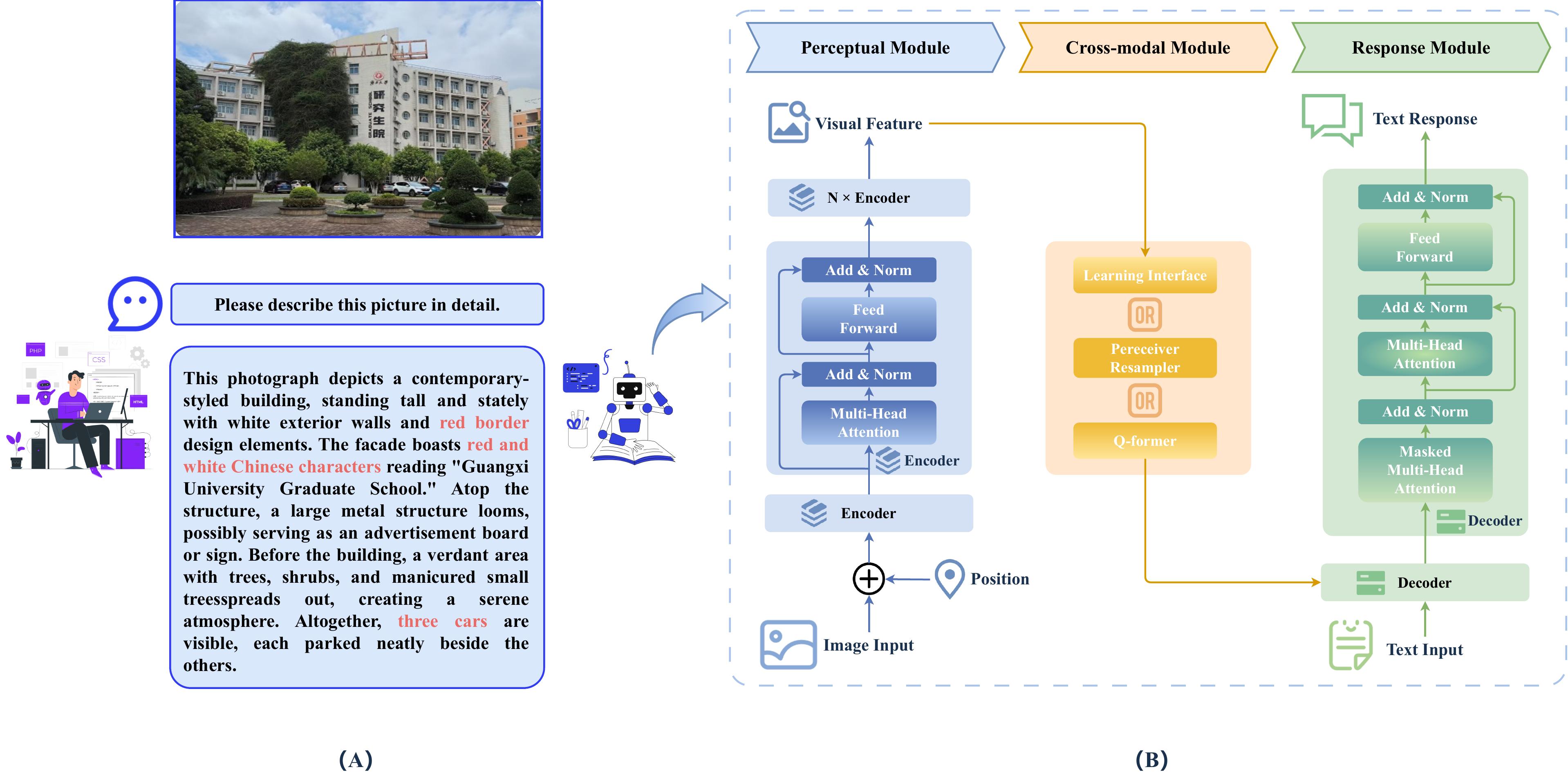}
\caption{(A) The examples of hallucinatory phenomena. The red font indicates the hallucinatory part of the LVLMs response. (B) The framework of LVLMs.}
\label{fig1}
\end{figure*}

\subsection{Structure of LVLMs}
LVLMs can be divided into three modules: perceptual module, cross-modal module and response module which is shown in Fig.~\ref{fig1}(B). Through the three modules, the visual information is extracted and mapped to the textual space. Further, the visual information and text information are combined  to get the final response. 

The perceptual module usually utilises Vision Transformer (ViT) \cite{13} or its variants \cite{64} to transform image into high-dimensional vector. Before input to ViT, the image is segmented into patches and added with positional information. As shown in Fig.~\ref{fig1}(B), the ViT is a encoder-only model which consists of N encoders. The multi-head attention of encoder is the core component of the Transformer model. It has powerful parallel computing capabilities and allows the model to create connections between different parts of the sequence.

Cross-modal module aims to bridge the modality gap between vision and language \cite{67}. Recently, cross-modal module in LVLMs adopts the structure such as learnable interface \cite{60,55}, Q-former \cite{52} and pereceiver resampler\cite{76}. The learnable interface maps visual information into textual space based on projection matrices. The Q-former bridges the modality gap by interacting visual information with text. The pereceiver resampler encodes visual features into text by using cross attention.

The response module acts as the brain of LVLMs. Therefore, it needs the powerful ability to process and analyse the visual and textual inputs to generate the final answer. The response module usually adopts LLMs such as Vicuna \cite{65}, LLaMa \cite{62}, Flan-PaLM \cite{66} and LLaMa2 \cite{63}. Both ViT and LLM are based on Transformer, but LLM is decoder-only structure. The masked multi-head attention of decoder adds the mask operation. Therefore, the LLM can not utilize the ``future" information in the text generation which ensures the authenticity.

\subsection{Causes of Hallucination}
Hallucinations in LVLMs originate from multiple components, including the perceptual module, the cross-modal module and the response module. Therefore, in order to better correct and mitigate hallucinations, we attribute the main causes of hallucinations as follows:

\subsubsection{Multi-modal Fusion Challenges}
Visual and textual modalities have distinct characteristics, creating a natural heterogeneity in their feature distributions and semantic representations \cite{liang2024foundations}. The modality gap biases the response module's interpretation of visual inputs, leading to the generation of hallucinated responses. For example, as shown in Fig.~\ref{fig1}(A), the red and white object is actually a sign, not a Chinese character. Due to the presence of the modality gap, the response module incorrectly describes it as a ``red and white Chinese character".
\subsubsection{Low-quality Data}
Training with cross-entropy loss inherently encourages the model to fit the training distribution \cite{yang2026detecting}. Therefore, LVLMs  tend to learn the statistical patterns from the dataset to generate responses, which are similar to the training data. As LVLMs require extensive data for training, training corpora are frequently augmented with synthetic data from other models. Although these data are manually cleaned after generation, noise and hallucinations often remain within these datasets. Consequently, when LVLMs train on such noisy data, they naturally inherit and reproduce hallucinations.
\subsubsection{Inherent Model Uncertainty}
The remarkable performance of LVLMs is mainly attributed to the use of LLMs as their backbone \cite{zhang2023instruction}. However, LLMs are easily to generate hallucinations. In addition, LLMs have acquired rich parametric knowledge. When this parametric knowledge is wrong or conflicts with the visual input, it leads to hallucinations. Moreover, the randomness of the available decoding strategies may also trigger hallucinations. Several specific phenomena occurring during the decoding process are closely related to hallucinations \cite{he2025survey}.

\begin{figure*}
\centering
\includegraphics[width=6in]{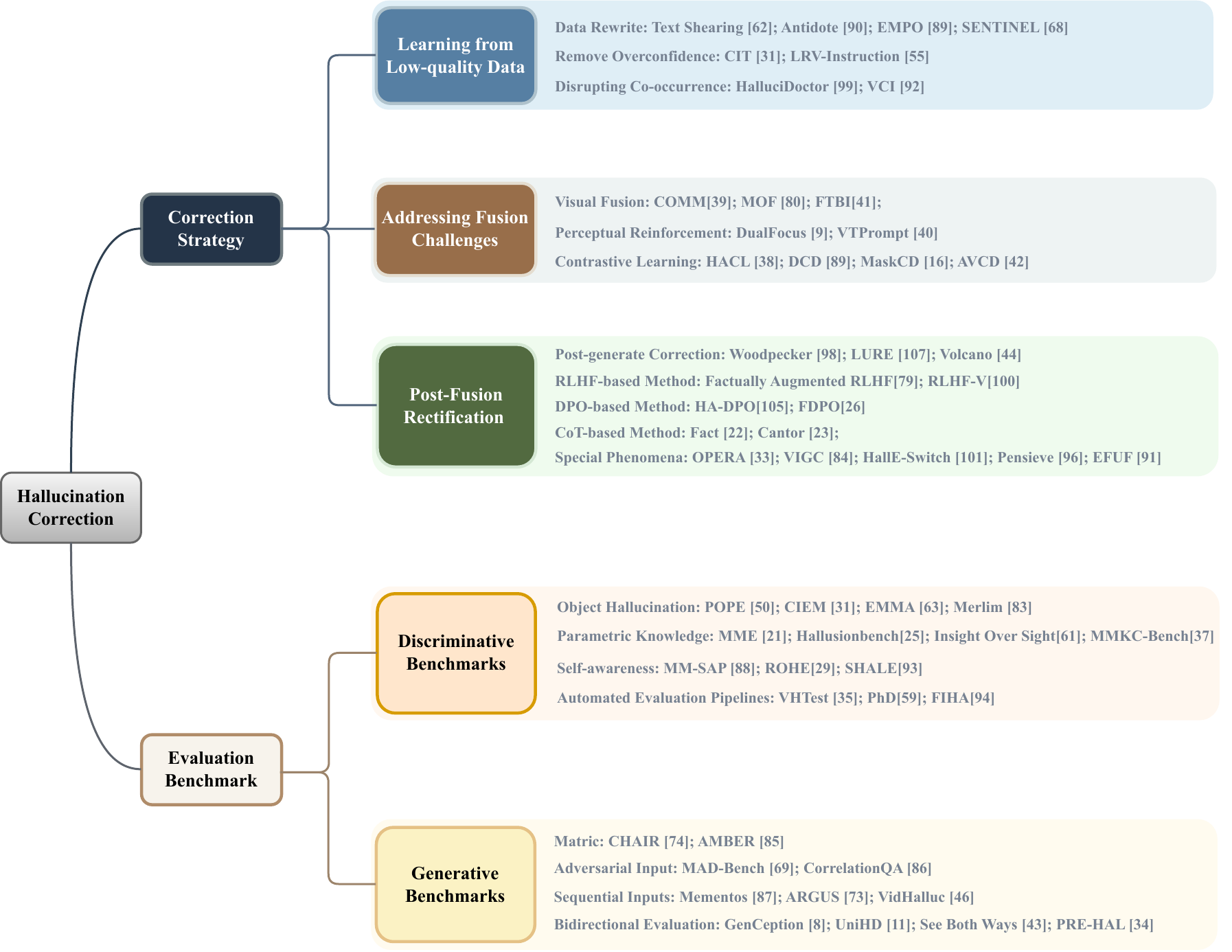}
\caption{A summary of hallucination correction methods across correction strategy and evaluation benchmark.} \label{hallucination-correction}
\end{figure*}

\section{Corrections of Hallucination}
Since Multimodal Foundation Models rely heavily on massive web-crawled data which is inherently noisy, a significant portion of mitigation methods aim to clean, rewrite, or re-weight data.
In this section, we summarize the core ideas of recent hallucination correction and mitigation works. Specifically, we consider the relationship between the recent studies and the causes of the hallucinations. We have categorized recent works into three classes: \textbf{Learning from Low-quality Data}, \textbf{Addressing Fusion Challenges} and \textbf{Post-Fusion Rectification}, which is shown in Fig.~\ref{hallucination-correction}. In addition, the details of all methods are summarized in Table.~\ref{tab:table1}.

\renewcommand{\arraystretch}{1.25} 

\begin{longtable}{C{2.0cm} L{5.0cm} C{1.5cm} L{6.5cm}}

\caption{A Comprehensive Overview of Correction and Refinement Methods for Multimodal Large Language Models (MLLMs). The methods are grouped by training requirements: the upper section lists \textbf{Training-Free} methods (highlighted in \colorbox{gray!60!blue!10!white}{Blue}), and the lower section lists \textbf{Training-Required} methods (highlighted in \colorbox{gray!60!red!10!white}{Red}).} \label{tab:table1} \\
\toprule[1pt] 
\textbf{Method} & \textbf{Targeted Issues/Goals} & \textbf{Training Required} & \textbf{Source/Repository} \\
\midrule
\endfirsthead

\caption[]{A Comprehensive Overview of Correction and Refinement Methods (Continued).} \\
\toprule[1pt] 
\textbf{Method} & \textbf{Targeted Issues/Goals} & \textbf{Training Required} & \textbf{Source/Repository} \\
\midrule
\endhead

\multicolumn{4}{r}{\textit{Continued on next page...}} \\
\endfoot

\bottomrule[1pt] 
\endlastfoot



\rowcolor{gray!60!blue!10!white}
Text Shearing & Noise data; Mismatched data; Long-tail distribution & Free & https://github.com/lyq312318224/MLLMs-Augmented \\ 

CIT & Object Hallucination; Over-confidence & Free & -- \\ 

\rowcolor{gray!60!blue!10!white}
LRV-Instruction & Object Hallucination; Over-confidence & Free & https://fuxiaoliu.github.io/LRV/ \\ 

HalluciDoctor & Object Hallucination & Free & https://github.com/Yuqifan1117/HalluciDoctor \\ 

\rowcolor{gray!60!blue!10!white}
VTPrompt & Visual Prompting; Textual Prompting & Free & https://github.com/jiangsongtao/VTprompt \\ 

Woodpecker & Object Hallucination & Free & https://github.com/BradyFU/Woodpecker \\ 

\rowcolor{gray!60!blue!10!white}
Cantor & Chain-of-Thought (CoT) & Free & https://ggg0919.github.io/cantor/ \\

OPERA & Knowledge Aggregation Pattern & Free & https://github.com/shikiw/OPERA \\ 

\rowcolor{gray!60!blue!10!white}
Pensieve & Mitigating Perception Module Errors & Free & https://github.com/DingchenYang99/Pensieve \\ 

VCI & Spurious Correlations; Causal Logic & Free & https://github.com/zhexu-ustc/Causal-HalBench \\

\rowcolor{gray!60!blue!10!white}
MAI & Object Hallucination (Head Intervention) & Free & -- \\

DCD & Object Hallucination; Contrastive Decoding & Free & -- \\

\rowcolor{gray!60!blue!10!white}
MaskCD & Object Hallucination; Image Heads & Free & https://github.com/Deng-Jingyuan/MaskCD \\

AVCD & Audio-Visual Hallucination; Modality Conflict & Free & https://github.com/kaistmm/AVCD \\

\midrule 


\rowcolor{gray!60!red!10!white}
COMM & Enhancing Visual Details & Need & -- \\

MOF & Enhancing Visual Details & Need & -- \\  

\rowcolor{gray!60!red!10!white}
DualFocus & Enhancing Visual Details & Need & https://github.com/InternLM/InternLM-XComposer/blob/main/projects/DualFocus \\ 

HACL & Object Hallucination & Need & -- \\ 

\rowcolor{gray!60!red!10!white}
LURE & Co-occurrence bias; Long-tail distribution & Need & https://github.com/YiyangZhou/LURE \\ 

Volcano & Iterative Self-Revision & Need & https://github.com/kaistAI/Volcano \\ 

\rowcolor{gray!60!red!10!white}
Factually Augmented RLHF & Aligning with Human Preferences & Need & https://llava-rlhf.github.io/ \\ 

RLHF-V & Aligning with Human Preferences & Need & https://rlhf-v.github.io/ \\ 

\rowcolor{gray!60!red!10!white}
HA-DPO & Aligning with Human Preferences & Need & -- \\ 

Fact & Chain-of-Thought (CoT) & Need & -- \\

\rowcolor{gray!60!red!10!white}
VIGC & Long-tail distribution & Need & https://opendatalab.github.io/VIGC/ \\ 

Halle-Switch & Parametric Knowledge Control & Need & https://github.com/bronyayang/HallE\_Switch \\ 

\rowcolor{gray!60!red!10!white}
EFUF & Text-Image Similarity Alignment & Need & -- \\ 

Antidote & Counterfactual Presupposition; Object Perception & Need & https://github.com/Wu0409/Antidote \\

\rowcolor{gray!60!red!10!white}
EMPO & Object Hallucination; Entity-Centric Preference & Need & -- \\

SENTINEL & Object Hallucination; Sentence-level Preference & Need & -- \\

\end{longtable}

\begin{figure*}
\centering
\includegraphics[width= \textwidth]{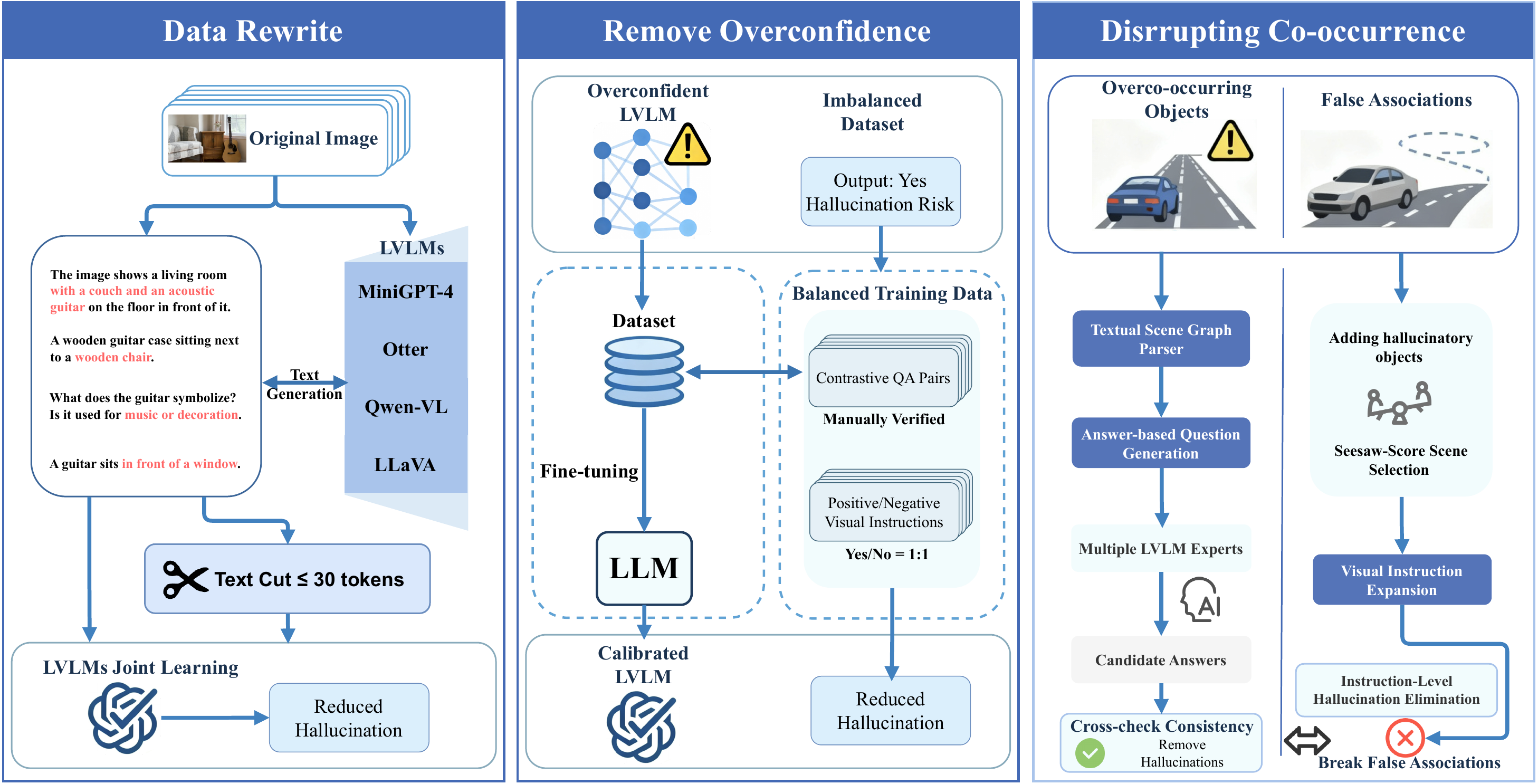}
\caption{Correction strategies of Learning from Low-quality Data.} \label{dataset-dehallucination}
\end{figure*}
\subsection{Learning from Low-quality Data}
LVLMs usually use instruction tuning to achieve powerful inference performance \cite{zhang2023instruction}. However, they often rely on high-quality and large-scale instruction datasets. In reality, it is difficult to construct high-quality instruction datasets even with the assistance of LLMs or LVLMs. Moreover, it is hard to create high-quality and large-scale datasets from scratch \cite{rzanny2025systematic}. Therefore, mitigating hallucinations within existing datasets presents a more practical and effective solution. In this section, we present recent work with three core ideas: \textbf{Data Rewrite}, \textbf{Remove Overconfidence} and \textbf{Disrupting Co-occurrence}, which is shown in Fig.~\ref{dataset-dehallucination}.


\subsubsection{Data Rewrite} 
Data rewrite aims to transform noisy, erroneous, or mismatched samples into high-quality training data, typically by leveraging the capabilities of LLMs or LVLMs. 
Early approaches focused on improving data diversity while controlling noise. For instance, Liu et al. \cite{8} generated diverse captions using multiple LVLMs (LLava-1.5\cite{55}, Otter \cite{56}, MiniGPT-4 \cite{57}). To eliminate hallucinations and style inconsistencies, they applied a text shearing strategy that retains only the semantically robust initial sentences. This approach effectively preserves main semantics and, through joint training, alleviates dataset toxicity.
Moving beyond general approaches, recent works utilize rewriting strategies to address specific hallucinatory triggers. To address the vulnerability of LVLMs in Counterfactual Presupposition Questions (CPQs), Wu et al. \cite{11095051} proposed the Antidote framework. They rewrote inquiries by incorporating factual priors, thereby guiding the model to self-correct and reject false premises. 
Similarly, to tackle modality misalignment of data, Wu et al. \cite{wu2025mitigating} introduced EMPO. They developed an entity-centric strategy to automatically synthesize high-quality preference data across image, instruction, and response dimensions. This approach resolves data scarcity and prioritizes image-text alignment over simple human preference matching.
Finally, to address the error propagation inherent in autoregressive generation, Peng et al. \cite{peng2025mitigating} proposed SENTINEL. This method constructs sentence-level preference pairs via detector cross-checking. It employs a context-aware preference loss (C-DPO) to enforce early intervention, effectively arresting the snowballing of hallucinations without relying on human annotations.

\subsubsection{Remove Overconfidence} If the dataset contains too many positive samples, it may lead to overconfidence (i.e. LVLMs respond ``Yes'' without any basis). To avoid overconfidence, Hu et al. \cite{10} proposed CIT to remove overconfidence by fine-tuning in a series of factual and contrastive question-answer (QA) pairs. These QA pairs are constructed by prompting chatGPT which contain balanced number of ``Yes'' and ``No'' answers. In QA pair, the questions focuses on hallucinatory scenes of objects existence, properties and inter-relationships. In addition, QA pairs are manually verified to ensure high quality. Similarly, Liu et al. \cite{9} constructed the LRV-Instruction by using GPT-4 \cite{15}, which contains a series of positive and negative visual instructions. Additionally, LRV-Instruction adds an examination of parametric knowledge in LVLMs by modifying the knowledge in the original instruction. By enforcing a balanced distribution of positive and negative samples during fine-tuning, both QA pairs and LRV-Instruction effectively reduce overconfidence and mitigate model hallucinations.


\subsubsection{Disrupting Co-occurrence} Since most of images in the dataset come from websites, it is inevitable that some objects such as "cars" and "roads" are frequently co-occurring. These co-occurrences affect the inference of LVLMs which leads to describe non-existent objects in responses. To address the co-occurring and hallucinatory objects in the dataset, Yu et al. \cite{7} proposed the HalluciDoctor framework based on the hallucination cross-checking paradigm and seesaw-based visual instruction expansion. As shown in Fig.~\ref{dataset-dehallucination}, the hallucination cross-checking paradigm is designed to find and remove hallucinations from instruction datasets. First, answer chunks are generated by using the textual scene graph parser \cite{107}. Then, answer-based questions are generated by chatGPT. Images and answer-based questions are input into multiple LVLM experts to generate candidate answers. Finally, the hallucinatory part of the instruction is identified and cleared by cross-checking the consistency between candidate answers and answer chunks. The seesaw-based visual instruction expansion aims to destroy the original false associations. The enhancement factor and inhibiting factor of the hallucinatory object are calculated to obtain the seesaw score, which is used to guide the tool model to integrate the hallucinatory object into irrelevant images and text. The enhancement factor $\mathcal{E}_i$ and inhibiting factor $\mathcal{I}_i$ are defined as follows:
\begin{equation}
\left.\mathcal{E}_i=\left\{\begin{array}{cc}\frac{n^*}{\max(n_i,1)},&\text{if }n_i\leq n^*\\1,&\text{if }n_i>n^*\end{array}\right.\right.
\end{equation}
\begin{equation}
\left.\mathcal{I}_i=\left\{\begin{array}{cc}\frac{m_i}{n^*},&\mathrm{if~}m_i\leq n^*\\1,&\mathrm{if~}m_i>n^*\end{array}\right.\right.
\end{equation}
where $o_h$ denotes hallucinatory object, and  $o_r $ denotes the object most relevant to $o_h $ in the context.  $n^\ast$ represents the number of their co-occurrences. $n_i$ denotes the number of co-occurrences of $o_h$ with other objects $o_i$. The smaller $n_i$ means less co-occurrence between $o_i$ and $o_h$, thus larger enhancement factor $\mathcal{E}_i$. $m_i$ denotes the number of co-occurrences of $o_r$ with other objects $o_i$. The inhibiting factor $\mathcal{I}_i$ is designed to ensure the reasonable of context. The smaller $m_i$ represents lower rationality, thus lower inhibiting factor $\mathcal{I}_i$. The seesaw score $\mathcal{S}_i$ is calculated based on enhancement factor and inhibiting factor, which is defined as follows: 
\begin{equation}
\mathcal{S}_i=\mathcal{E}_i*\mathcal{I}_i
\end{equation}
The seesaw score represents the object with least relevant to the hallucinated object $o$. The effect of destroying false association is achieved by integrating $o$ into the image which has the highest seesaw score. HalluciDoctor obtains high-quality instruction datasets by cleaning instruction-level and image-level hallucination. Meanwhile, HalluciDoctor is free for training which is a resource-friendly data cleaning framework.

While data-level manipulations like HalluciDoctor effectively mitigate statistical biases, recent advancements have focused on identifying and severing the underlying causal links within the model itself. 
Researchers have focused on addressing these spurious correlations through causal intervention and internal mechanism decoupling. Xu et al. \cite{xu2025causal} identified that LVLMs rely on non-causal paths established by object co-occurrence (e.g., inferring skis solely from a snowy background). They introduced a Visual Content Intervention (VCI) strategy to construct counterfactual scenarios, effectively disrupting these contextual biases and forcing the model to rely on actual visual evidence. 
Similarly, addressing the internal propagation of co-occurrence priors, Yang et al. \cite{yang2025mitigating} discovered that specific ``hallucination heads'' in the attention mechanism are responsible for over-attending to textual history rather than visual inputs. By adaptively deactivating these heads during decoding, they successfully disrupted the model's reliance on language-based co-occurrence patterns without retraining the entire model.

\begin{figure*}
\centering
\includegraphics[width= \textwidth]{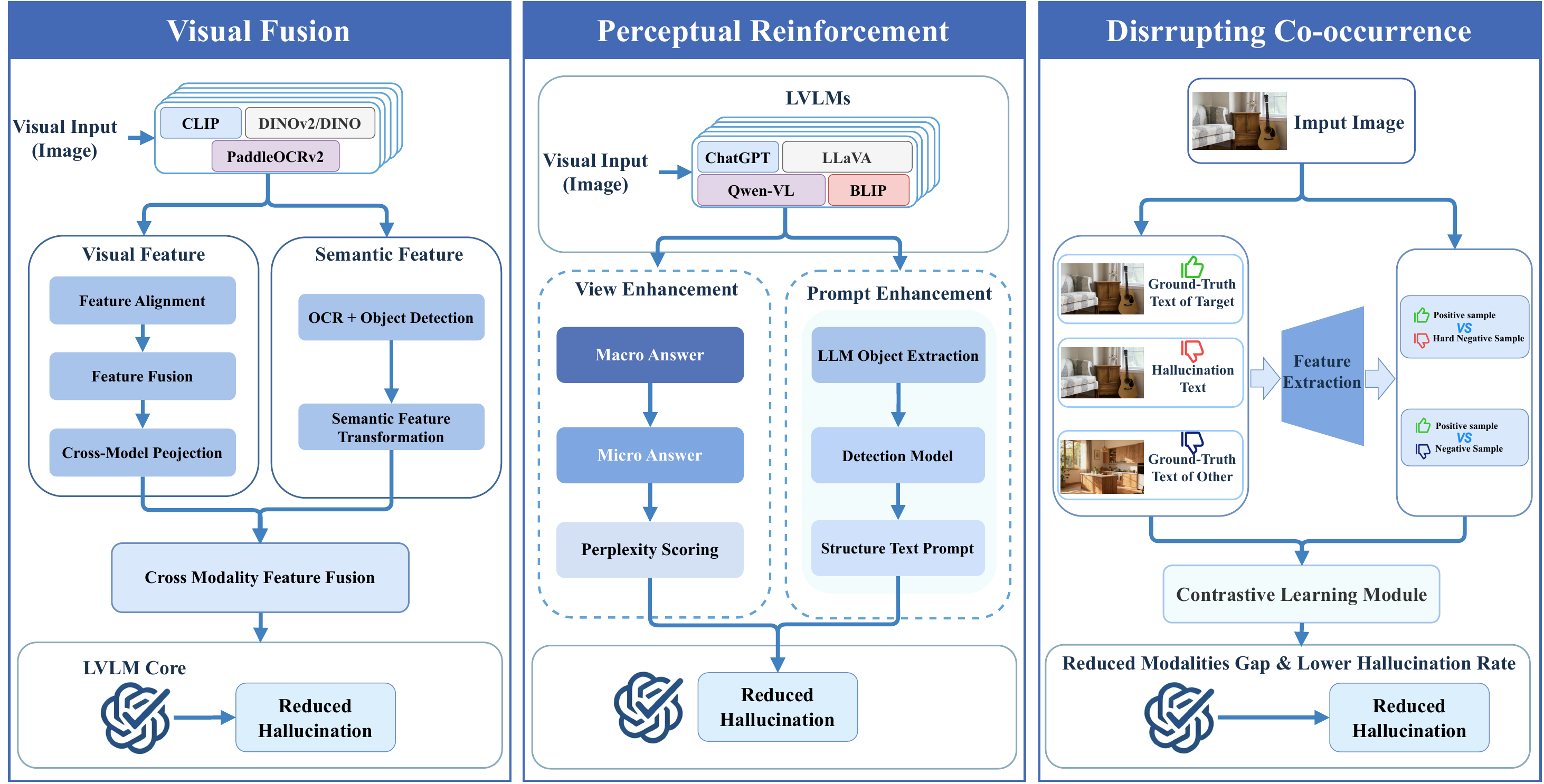}
\caption{Correction strategies of Addressing Fusion Challenges.} \label{Modalities-Gap}
\end{figure*}

\subsection{Addressing Fusion Challenges}
To overcome the fusion challenges originating from the heterogeneity of vision and language, these methods focus on better feature alignment and integration.
LVLMs often rely on the parametric knowledge within the response module to generate answers when the perception module fails to capture sufficient visual information. Consequently, hallucinations arise if this parametric knowledge conflicts with the ground-truth visual content. Furthermore, the cross-modal module serves as a critical bridge in LVLMs. If a misalignment persists between visual features and the textual space after mapping, it can induce biases in how the response module interprets visual information. Therefore, enhancing the capabilities of extracting and mapping visual information is essential for mitigating hallucinations. In this section, we categorize related works into \textbf{Visual Fusion}, \textbf{Perceptual Reinforcement}, and \textbf{Contrastive Learning}, which is shown in Fig. \ref{Modalities-Gap}.


\subsubsection{Visual Fusion} 
Different visual models possess distinct strengths in feature extraction. Fusing features from multiple encoders can effectively enhance the visual comprehension of LVLMs. 
Jiang et al. \cite{92} proposed COMM, a strategy to enhance visual comprehension by integrating CLIP and DINOv2. In this method, the feature spaces of different layers are aligned via a Linear-LayerNorm (LLN) module. Subsequently, multi-layer features are merged using LayerScale. Additionally, a Multilayer Perceptron (MLP) projects DINOv2 features into the CLIP feature space to ensure consistency between the two vision models. Finally, the fused features are projected into the text space via a linear layer to strengthen the LVLM's perception of visual details. 
Tong et al. \cite{12} proposed Mixture-of-Features (MoF) to combine features from CLIP and DINOv2, achieving richer visual understanding without additional training. 
Similarly, Jiao et al. \cite{22} proposed FTBI, utilizing DINO \cite{23} and PaddleOCRv2 \cite{24} to capture more comprehensive visual information. Specifically, object detection and Optical Character Recognition (OCR) results are extracted using DINO and PaddleOCRv2, respectively. These results are then transformed into text features through the LLM's embedding layer. Finally, both the transformed text features and the visual features extracted by CLIP are input into the LLM. 
These fusion strategies significantly improve the visual perceptual capability of LVLMs, thereby mitigating the generation of hallucinations.

\subsubsection{Perceptual Reinforcement} 
The image input for the perception module is typically restricted to a fixed resolution (e.g., $224\times 224$), which limits the LVLM's ability to capture fine-grained visual details. To address this, Cao et al. \cite{34} proposed DualFocus, a framework that generates responses from both macro and micro perspectives. Specifically, DualFocus utilizes the original image $I_o$ to generate a macro answer. For the microscopic perspective, it employs the LVLM to identify sub-region coordinates $\hat{box}$ relevant to the user question $Q_1$, from which a sub-region image $I_s$ is cropped. Simultaneously, a refined question $Q_2$ is derived by adapting $Q_1$ with specific prompts. Subsequently, $I_o$, $I_s$, $Q_1$, and $Q_2$ are fed into the LVLM to produce a micro answer. The credibility of both answers is assessed via perplexity scores, with the lower-perplexity response selected as the final output. This mechanism significantly enhances the visual perception capabilities of LVLMs.
	
Beyond resolution, object detection models can provide explicit visual information, such as object counts, locations, and attributes. Leveraging this, Jiang et al. \cite{93} proposed VTPrompt to enhance LVLM perception. VTPrompt first employs ChatGPT to extract key objects from user queries. It then utilizes a detection model (SPHINX \cite{94}) to annotate these main objects within the image, thereby providing precise location information. Prior to response generation, VTPrompt applies structured textual prompts for query transformation. This guides the LVLM to generate a ``visual chain of thought'' by leveraging the annotated visual cues. Finally, the LVLM generates responses based on the marked images and processed queries. Consequently, VTPrompt effectively improves the interpretability and perceptual grounding of the LVLM.


\subsubsection{Contrastive Learning}
The core of contrastive learning is to extract robust representations by maximizing the similarity between positive pairs while minimizing it for negative ones. In the context of LVLMs, this paradigm addresses the distributional discrepancy between factual and hallucinatory responses. From the training perspective, Liu et al. \cite{14} proposed HACL to mitigate hallucinations via hard-negative mining. It utilizes ground-truth text as positive samples, hallucinatory text as hard negative samples, and captions from unrelated images as soft negative samples. By enlarging the distance between visual features and hallucinatory textual features, HACL effectively reduces the modality gap and prevents the model from generating non-existent content during the alignment phase.

More recently, the focus has shifted towards Inference-Time Contrastive Decoding, which mitigates hallucinations by dynamically contrasting the output logits of a standard model against a hallucination-prone baseline. To enhance the robustness of this approach, Dual Contrastive Decoding (DCD) \cite{wu2025mitigating} was proposed. This method introduces a dual-stream contrastive mechanism that simultaneously contrasts the LVLM's predictions against both a text-only baseline and a perturbed-visual baseline, effectively filtering out language priors and visual misinterpretations. Focusing on internal interference, MaskCD \cite{deng2025maskcd} innovates by identifying and masking specific ``image heads'' within the attention mechanism that are responsible for visual processing. By contrasting the logits of the full model with this visually-impaired version, MaskCD accurately isolates and amplifies visual evidence without requiring computationally expensive external image perturbations. Furthermore, the efficacy of this paradigm extends beyond vision-language tasks. AVCD \cite{jung2025avcd} applies contrastive decoding to Audio-Visual Large Language Models, demonstrating that contrasting multimodal inputs against unimodal baselines can effectively mitigate hallucinations across diverse modality combinations.

\begin{figure*}
\centering
\includegraphics[width= \textwidth]{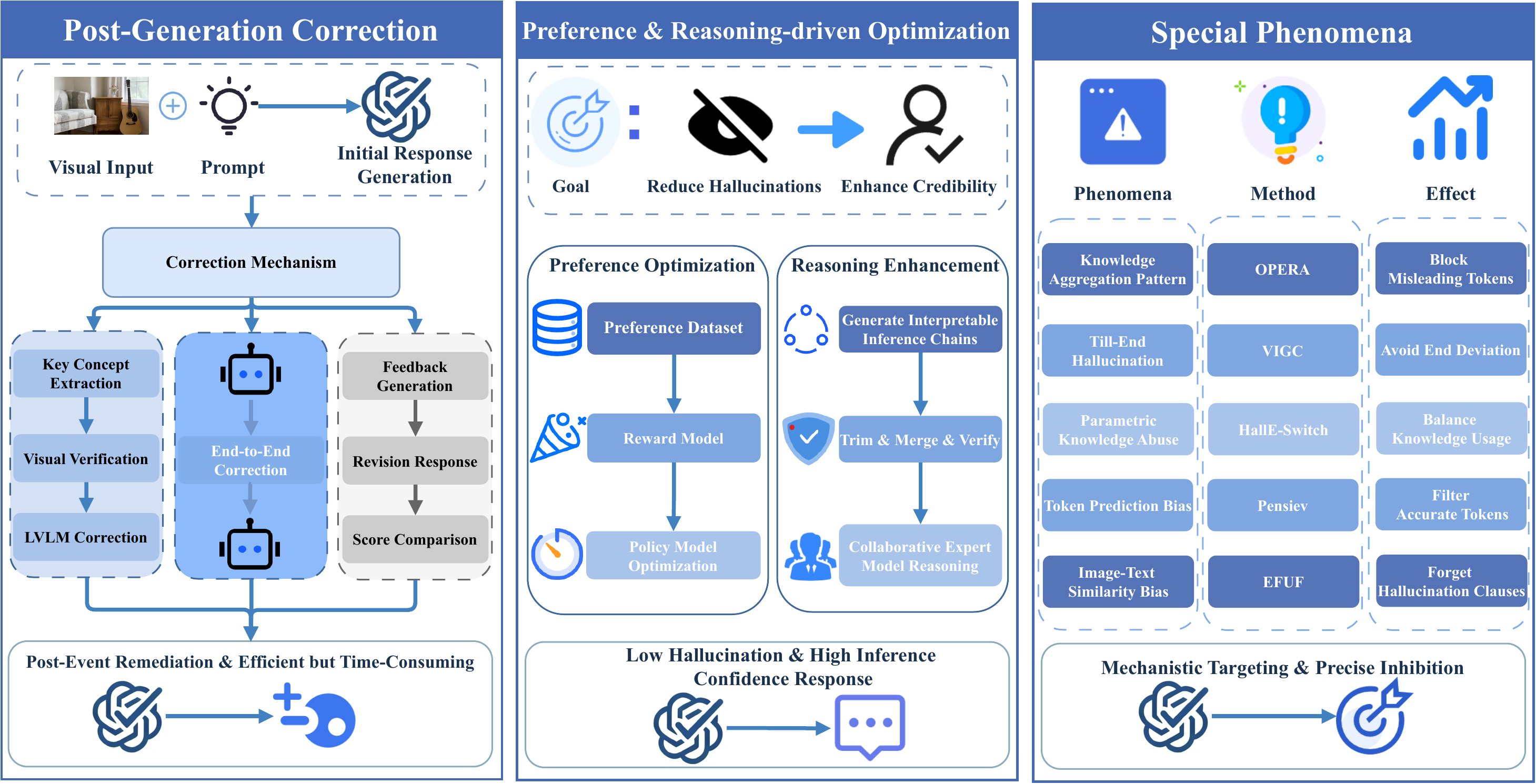}
\caption{Correction strategies of Post-Fusion Rectification.} \label{Output-Correction}
\end{figure*}

\subsection{Post-Fusion Rectification}
	In hallucination correction, correcting the hallucinatory response to an accurate response is the most straightforward approach. Changing the output preference of LVLMs can also mitigate hallucinations. In addition, hallucinations are closely related to many phenomena in the decoding process of LVLMs. Analyzing these phenomena can help to understand the generation mechanism of hallucinations and mitigate the generation of hallucinations. In this section, related works are classified into \textbf{Post-Generation Correction}, \textbf{RLHF-based Methods}, \textbf{DPO-based Methods}, \textbf{CoT-based Methods}, and \textbf{Special Phenomena}, which is shown in Fig. \ref{Output-Correction}.

\subsubsection{Post-Generation Correction}
To mitigate hallucinations, a direct approach is to perform post hoc remediation by detecting and rectifying errors in the generated response. Yin et al. \cite{16} proposed Woodpecker to achieve this. In this framework, an LLM first extracts key concepts from the initial response and constructs questions regarding the main objects. Subsequently, visual validation is performed using answers provided by an open-set object detector \cite{17} and a VQA model \cite{6}. Finally, the LLM utilizes these QA pairs as guidance to rectify hallucinations.

To streamline this process, Zhou et al. \cite{20} trained a single LVLM hallucination revisor (LURE) instead of relying on multiple expert models. LURE takes images and hallucinatory descriptions as input, training with correct descriptions as the target output. However, this method remains sensitive to object co-occurrence, which may inadvertently induce hallucinations.

Furthermore, LVLMs can reduce hallucinations by iteratively refining their own responses. Lee et al. \cite{98} proposed Volcano, a method implementing this iterative correction mechanism. Volcano initiates by inputting the image and question to generate an initial response ($Response_{I}$). Next, the LVLM is prompted to generate self-feedback on this response. Based on this feedback, the model produces a revised response ($Response_{R}$). Finally, the LVLM calculates confidence scores for both the initial ($score_{I}$) and revised ($score_{R}$) responses. The model outputs $Response_{I}$ if $score_{I} > score_{R}$; otherwise, the iteration continues (or accepts the revision). While post-generation correction effectively mitigates hallucinations, it typically incurs higher computational latency due to the additional inference steps.


\subsubsection{RLHF-based Methods}
Reinforcement Learning from Human Feedback (RLHF) \cite{40,41,42} aims to optimize model behavior by utilizing human feedback as a reward signal. Factually augmented RLHF (Fact-RLHF) \cite{39} pioneered the application of RLHF in the multi-modal domain. This method involves three training stages. In the first stage, the instruction dataset is used to supervised fine-tuning the LVLM, yielding a policy model. In the second stage, Fact-RLHF constructs a hallucination-aware human preference dataset, upon which a reward model is trained to provide accurate reward signals. In the third stage, the policy model is optimized by maximizing these reward signals. To prevent reward hacking, Fact-RLHF additionally incorporates ground-truth information to calibrate the reward mechanism.

To enhance efficiency, RLHF-V \cite{19} eliminates the need for separate reward model training by employing the Dense Direct Preference Optimization (DDPO) strategy. RLHF-V first constructs a dataset containing segment-level, fine-grained correctional human feedback. The optimization process then replaces the explicit reward model with a policy model and a frozen reference model. The training objective is to increase the relative log-probability of human-preferred segments compared to the rejected segments. Unlike PPO-based methods, only the policy model is updated during this process.
Furthermore, to fully leverage segment-level information, RLHF-V calculates the overall response score by weighting fine-grained segments. It applies specific weighting hyperparameters to corrected fragments to distinguish them from uncorrected segments. By optimizing the LVLM with these fine-grained preferences, the model achieves a better understanding of human judgments regarding hallucinations, thereby significantly improving its credibility.



\subsubsection{DPO-based Methods}
Direct Preference Optimization (DPO) \cite{28} aims to directly optimize the policy model from preference data, thereby enhancing the efficiency of reinforcement learning by bypassing the reward modeling step. Based on DPO, Zhao et al. \cite{27} proposed Hallucination-Aware DPO (HA-DPO). To train this model, they utilized a style-consistent hallucination dataset containing images, text prompts, positive responses , and negative (hallucinatory) responses. The optimization process directly adjusts the policy model against a reference model using these pairs. This approach effectively encourages the LVLM to assign higher probability to positive responses while rejecting negative ones.

To address granularity issues, Gunjal et al. \cite{29} employed a DPO variant: Fine-grained Direct Preference Optimization (FDPO). FDPO leverages the fine-grained M-HalDetect dataset to provide precise preference signals. Unlike simpler methods, this dataset features segment-level annotations, categorizing segments as accurate, inaccurate, or analytical. The optimization process utilizes these segment-level annotations to guide the model, enabling it to distinguish between positive, negative, and neutral signals at a sub-sentence level. Finally, rejection sampling is employed during inference to select the response with the lowest hallucination risk.


\subsubsection{CoT-based Methods}
Chain of Thought (CoT) aims to enhance the reasoning capabilities of models by generating a reasoning process prior to the final answer, thereby facilitating better comprehension and problem-solving. However, reasoning in LVLMs may stem from spurious correlations derived from their powerful representational capabilities, rather than genuine interpretability \cite{88}. 

To address this interpretability gap, Gao et al. \cite{89} proposed the Fact method. This approach utilizes code generation models to produce interpretable code snippets that derive the correct answer. These snippets are transformed into a CoT via pruning, merging, and bridging operations, followed by a transferability verification step to filter irrelevant components. Finally, the LVLM is jointly trained with the generated CoT and corresponding labels to mitigate hallucinations.

Furthermore, Gao et al. \cite{99} observed that while LVLMs capture higher-level visual semantics compared to specialized tools (e.g., detectors and OCR engines), they can also serve as orchestrators for these expert models. To leverage this, they proposed the Cantor method, which enhances visual reasoning by guiding the LVLM to assume multiple roles for reasoning, decision-making, and execution. The inference process comprises two phases: decision generation and execution. In the decision generation phase, Cantor prompts the LVLM to reason about the query and decompose it into tasks for expert models. In the execution phase, the LVLM is prompted to function as these distinct experts, completing the assigned sub-tasks. Finally, the LVLM serves as an information integration expert to aggregate all sub-task results and determine the final answer.

\subsubsection{Special Phenomena}
Specific phenomena and decoding patterns are intrinsically linked to hallucinations in LVLMs. For example, Huang et al. \cite{18} identified the ``knowledge aggregation pattern,'' where certain summary tokens contain limited information yet disproportionately guide subsequent generation. To address this, they proposed OPERA, utilizing an over-trust penalty and a retrospection-allocation strategy. 

In the over-trust penalty component, OPERA analyzes self-attention weights within a localized window. It computes a column-wise score vector by zeroing out the upper triangles of the attention weights. The peak value in this vector indicates a knowledge aggregation pattern. During decoding, this score is combined with logits to penalize such aggregation. However, since penalties alone may not fully eliminate hallucinations, the retrospection-allocation strategy initiates a fallback mechanism. This mechanism triggers the model to re-predict the summary token whenever the pattern frequency exceeds a threshold across decoding rounds.

LVLMs generate ``tail-end hallucination'' at the end of a response when they prioritize language priors (answer tendencies) over visual cues. To address it, Wang et al. \cite{21} proposed the VIGC method to mitigate this via iterative generation. VIGC first generates an initial sentence based on the prompt. In subsequent iterations, the model produces the next sentence by conditioning on the instruction, the question, and the preceding text. This cycle continues until the model outputs a termination symbol. Finally, VIGC concatenates all generated segments to form the complete response.

During training, if the visual input conflicts with the ground truth, LVLMs often resort to ``guessing'' based on textual associations, forming parametric knowledge. Zhai et al. \cite{46} identified this parametric knowledge as a source of hallucination. However, since this knowledge also underpins the model's generative capabilities, it cannot be entirely discarded. Consequently, they introduced HallE-Switch to regulate parametric knowledge via a control parameter. This parameter scales the output of a learnable projector that transforms generic word embeddings into object-sensitive space. During training, the parameter is toggled between +1 (enabling parametric knowledge) and -1 (disabling it). During inference, users can fine-tune this parameter within the range of [-1, +1] to balance creativity and faithfulness.

In the decoding phase, Yang et al. \cite{95} proposed Pensieve, a method to differentiate between accurate and inaccurate candidate tokens by analyzing the impact of visual inputs. Pensieve introduces perturbed inputs, including similar images and Gaussian noise. During the decoding of a specific token, the model predicts probabilities conditioned on the original text and these varied visual inputs. The method then calculates a reference value based on the confidence scores across the original, similar, and noise images. Accurate tokens typically exhibit high variance between the original and similar images, as they are grounded in specific visual details present only in the original image. By prioritizing these tokens, Pensieve effectively mitigates hallucinations.

Xing et al. \cite{35} proposed an Efficient Fine-grained Unlearning Framework (EFUF), premised on the ability of CLIP scores to distinguish between hallucinatory and factual segments. EFUF constructs a fine-grained dataset comprising positive sub-sentences, negative (hallucinatory) sub-sentences, and sentence-level responses. It employs gradient ascent on negative sub-sentences to unlearn hallucinations. Specifically, the total loss combines: (1) a negative loss (inverted cross-entropy) for hallucinatory segments, (2) a positive loss for factual segments, and (3) a sentence-level loss to preserve text generation fluency. Weighted summation of these components allows EFUF to suppress hallucinatory content while maintaining coherence.

\begin{table*}[!t]
\centering
\small 
\caption{Overview of Evaluation Benchmarks and Examination Scenes. The table indicates whether the benchmark implementation code is publicly available.}
\label{tab:table2}
\renewcommand{\arraystretch}{1.25} 

\begin{tabular}{L{2.0cm} L{5.0cm} C{1.5cm} L{6.0cm}}
\toprule[1pt]
\textbf{Benchmark} & \textbf{Examination Scene} & \textbf{Code Available} & \textbf{Source / Repository} \\
\midrule

\rowcolor{rowblue}
POPE & Object Exists & \makecell{\xmarkcolor} & -- \\ 

CIEM & Object Exists; Properties; Actions & \makecell{\xmarkcolor} & -- \\ 

\rowcolor{rowblue}
EMMA & Object Exists; Properties; Actions; Placement & \makecell{\xmarkcolor} & -- \\

Merlim & Object recognition; Inter-object relations; Object counting & \makecell{\checkmarkcolor} & https://github.com/ojedaf/MERLIM \\ 

\rowcolor{rowblue}
MME & Object Exists; Properties; Knowledge resource & \makecell{\xmarkcolor} & -- \\ 

Hallusionbench & Application of parametric knowledge & \makecell{\checkmarkcolor} & https://github.com/tianyilab/HallusionBench \\ 

\rowcolor{rowblue}
MM-SAP & Self-awareness & \makecell{\checkmarkcolor} & https://github.com/YHWmz/MM-SAP \\ 

VHTest & Object Exists; Properties; Actions; Placement & \makecell{\checkmarkcolor} & https://github.com/wenhuang2000/VHTest \\ 

\rowcolor{rowblue}
CHAIR & Object Exists & \makecell{\xmarkcolor} & -- \\ 

AMBER & Object Exists; Properties; Relation; Inter-object relations & \makecell{\checkmarkcolor} & https://github.com/junyangwang0410/AMBER \\
 
\rowcolor{rowblue}
MAD-Bench & Fraudulent input & \makecell{\xmarkcolor} & -- \\ 

CorrelationQA & Fraudulent input & \makecell{\checkmarkcolor} & https://github.com/MasaiahHan/CorrelationQA \\ 

\rowcolor{rowblue}
GenCeption & Semantic consistency & \makecell{\xmarkcolor} & -- \\ 

Mementos & Dynamic inference & \makecell{\checkmarkcolor} & https://github.com/umd-huanglab/Mementos \\ 

\rowcolor{rowblue}
UniHD & Object Exists; Properties; Scenes; Knowledge resources & \makecell{\xmarkcolor} & -- \\ 

ROHE & Object Exists (Removed Objects) & \makecell{\xmarkcolor} & -- \\ 

\rowcolor{rowblue}
MMKC-Bench & Multimodal Knowledge Conflict & \makecell{\checkmarkcolor} & https://github.com/MLLMKCBENCH/MLLMKC \\ 

ConflictVis & Vision-Knowledge Conflicts & \makecell{\xmarkcolor} & -- \\ 

\rowcolor{rowblue}
PhD & Visual Hallucination (Task/Mode specific) & \makecell{\checkmarkcolor} & https://github.com/jiazhen-code/PhD \\ 

FIHA & Fine-grained Hallucinations; Scene Graphs & \makecell{\checkmarkcolor} & https://github.com/confidentzzzs/FIHA \\ 

\rowcolor{rowblue}
Bi-Eval & Image Captioning (Human vs. Model) & \makecell{\xmarkcolor} & -- \\ 

PRE-HAL & Hallucination Detection; Evidential Conflict & \makecell{\checkmarkcolor} & https://github.com/HT86159/Evidential-Conflict \\ 

\rowcolor{rowblue}
ARGUS & Video Hallucination; Omission & \makecell{\checkmarkcolor} & https://github.com/JARVVVIS/argus \\ 

VIDHALLUC & Video Temporal Hallucination & \makecell{\checkmarkcolor} & https://people-robots.github.io/vidhalluc \\

\rowcolor{rowblue}
SHALE & Factuality Hallucinations & \makecell{\checkmarkcolor} &  https://github.com/BeiiiY/SHALE \\

\bottomrule[1pt]
\end{tabular}
\end{table*}

\section{Evaluations of Hallucination}
Hallucination evaluation benchmarks are generally categorized into two types: discriminative and generative. Discriminative benchmarks assess LVLMs using a series of binary (Yes/No) questions. In contrast, generative benchmarks involve extracting key elements (e.g., objects) from the LVLM's response and comparing them against the ground truth. The specific evaluation scenarios and code repositories for each benchmark are presented in Table~\ref{tab:table2}.

\begin{figure*}
\centering
\includegraphics[width= \textwidth]{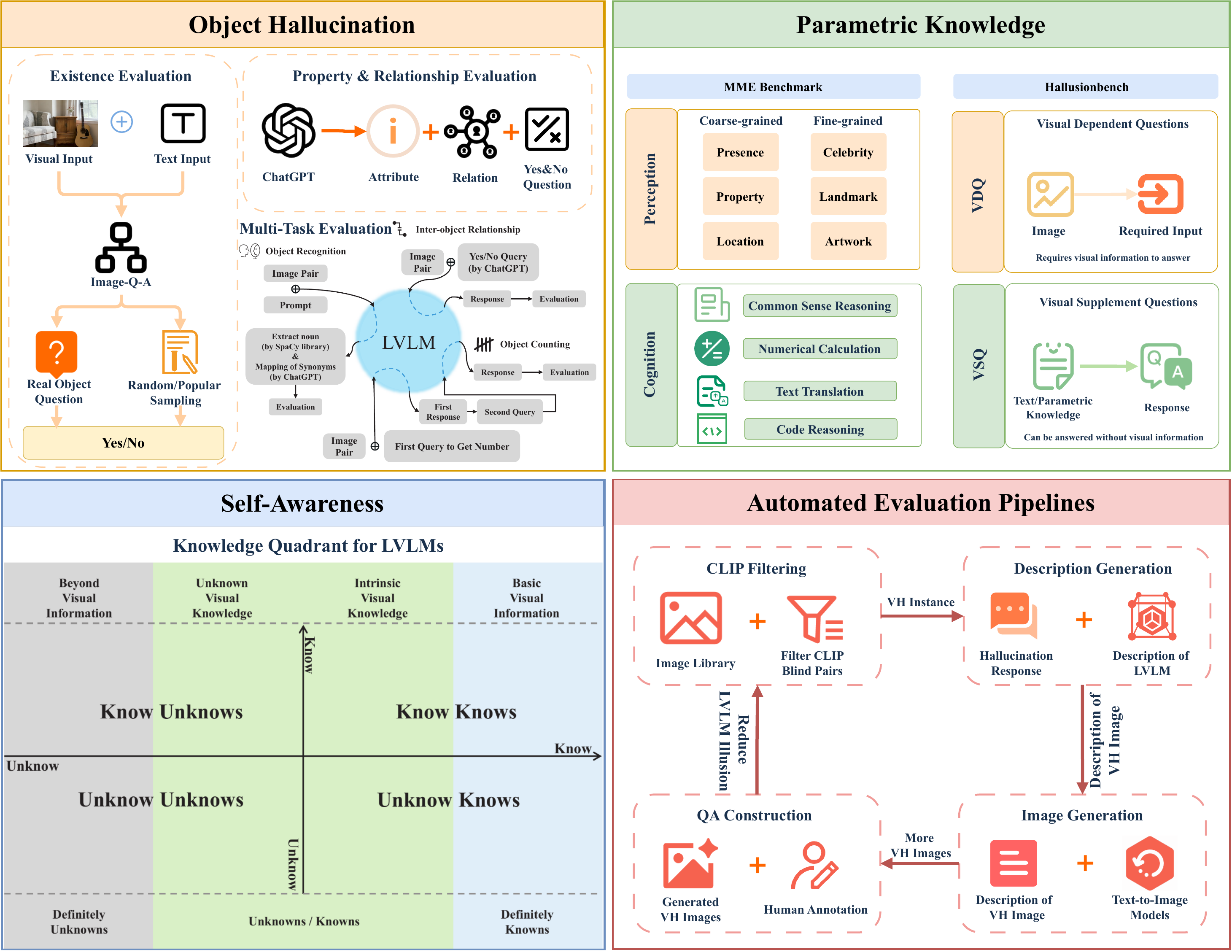}
\caption{Discriminative benchmarks} \label{Discriminative-benchmarks}
\end{figure*}

\subsection{Discriminative benchmarks}
Discriminative benchmarks assess the hallucination rates of LVLMs by probing their ability to distinguish factual visual content from fabricated information. Typically employing binary or multiple-choice formats, these benchmarks quantify the model's alignment with visual reality and established knowledge. As shown in Fig. \ref{Discriminative-benchmarks}, this section categorizes these evaluation paradigms into four primary dimensions. First, \textbf{Object Hallucination} focuses on verifying fine-grained visual details, including object existence, attributes, and inter-relationships. Second, \textbf{Parametric Knowledge} examines the conflicts between visual evidence and the model's internal priors, highlighting the reliance on pre-trained knowledge versus visual perception. Third, \textbf{Self-awareness} evaluates the model's capacity to recognize the boundaries of its capabilities, specifically its ability to identify object absence or refuse unanswerable queries. Finally, \textbf{Automated Evaluation Pipelines} address the scalability challenge by leveraging advanced algorithms to synthesize diverse and adversarial evaluation instances without manual intervention.

\subsubsection{Object Hallucination} 
Object hallucination refers to the phenomenon where LVLMs generate descriptions of non-existent objects, or incorrect object properties, behaviors, and inter-relationships. To evaluate object existence hallucinations, Li et al. \cite{26} proposed Polling-based Object Probing Evaluation (POPE). POPE constructs evaluation triples—comprising an image, a question, and a binary answer—derived from image caption datasets. For positive samples (Answer: ``Yes''), queries target objects present in the ground truth. For negative samples (Answer: ``No''), objects are selected using three strategies: random, popular, and adversarial sampling. Random sampling selects objects absent from the current image. Popular sampling selects the top-$k$ most frequent objects in the dataset. Adversarial sampling selects the top-$k$ objects that most frequently co-occur with the ground-truth objects yet are absent in the image.

Beyond coarse-grained existence, evaluation metrics have been extended to fine-grained attributes like properties and relationships. Hu et al. \cite{10} proposed the Contrastive Instruction Evaluation Method (CIEM), utilizing ChatGPT to generate binary questions regarding object existence, properties, and inter-relationships based on captions. Evaluation metrics include accuracy, precision, recall, specificity, and F1-score. Similarly, Lu et al. \cite{30} proposed the Evaluation and Mitigation of Multimodal Agnosia (EMMA) framework, which constructs benchmarks using multiple-choice questions (MCQs). EMMA employs template-based generation, filling placeholders with ground-truth data, while distractor options are generated via a thesaurus and manually verified for quality.

Villa et al. \cite{96} introduced the Merlim framework, encompassing three tasks: object recognition, relationship understanding, and counting. For object recognition, Merlim prompts the LVLM to list all visible objects, extracting nouns via spaCy \cite{109} for comparison with ground truth. For relationship understanding, it employs ChatGPT to generate two relationship sets: a ``random set'' containing absurd relations (e.g., ``Does a clock have wheels?'') and a ``curated set'' containing logical relations requiring visual verification (e.g., ``Are there drops of water on the mirror?''). In object counting, Merlim first prompts the model for a count (``How many...'') and subsequently verifies consistency via a follow-up confirmation question (``Is there [number] [object]?''). Furthermore, Merlim leverages image inpainting \cite{97} to create counterfactual images by removing objects. By comparing performance on original versus edited images, the framework identifies hallucinations driven by language priors rather than visual evidence.

\subsubsection{Parametric Knowledge} 
The rich parametric knowledge embedded in LVLMs is a double-edged sword: while it empowers cognitive capabilities, it is also closely related to hallucination generation. Consequently, the reliance on parametric knowledge cannot be adequately assessed solely by evaluating object existence. To comprehensively evaluate these cognitive aspects, Fu et al. \cite{33} proposed the MME benchmark. MME divides the evaluation into perceptual and cognitive abilities. Perceptual evaluation includes coarse-grained recognition (e.g., object existence, properties, and positions) and fine-grained recognition, which tests the model's internal knowledge regarding specific domains such as movie posters, celebrities, landmarks, and artworks. Cognition evaluation spans four tasks: commonsense reasoning, numerical calculation, text translation, and code reasoning. All instructions in MME are manually designed to ensure high quality. 
Similarly, Guan et al. \cite{43} introduced HallusionBench, a manually curated benchmark featuring two question types: Visual Dependent Questions (VDQ) and Visual Supplement Questions (VSQ). VDQs require visual information to be answered correctly, whereas VSQs rely on parametric knowledge and can be answered without visual input.

Recent studies have further investigated the specific conflicts between visual evidence and internal parametric knowledge. To explore commonsense-level vision-knowledge conflicts, \citet{liu2024insight} introduced an automated framework and a diagnostic benchmark comprising 1,122 high-quality QA pairs. Their study revealed that LVLMs exhibit a significant over-reliance on parametric knowledge, ignoring contradictory visual information in approximately 20\% of queries, particularly in Yes-No and action-related problems. 
Expanding this scope to Retrieval-Augmented Generation (RAG) scenarios, \citet{jia2025benchmarking} proposed MMKC-Bench to evaluate multimodal knowledge conflicts where external context contradicts internal memory. They constructed a dataset encompassing 1,573 knowledge instances across context-memory and inter-context conflict scenarios. Their findings indicate that while current models can recognize conflicts, they tend to stubbornly favor their internal parametric knowledge over external retrieved evidence, leading to unreliable outputs.

\subsubsection{Self-awareness} 
Self-awareness in LVLMs refers to the model's ability to recognize the boundaries of its capabilities, enabling it to distinguish between answerable and unanswerable queries to avoid hallucination. Post-instruction tuning, LVLMs are expected to not only master basic visual concepts (e.g., attributes, shapes, colors) but also to critically assess whether the visual evidence supports a definitive answer. 

Based on this premise, Wang et al. \cite{47} proposed a Knowledge Quadrant for LVLMs and constructed the MM-SAP benchmark. MM-SAP comprises three subsets: BasicVisQA, KnowVisQA, and BeyondVisQA. As shown in Fig.~\ref{Discriminative-benchmarks}, BasicVisQA corresponds to the ``Known Knowns'' (blue quadrant), assessing mastery of basic visual concepts. KnowVisQA corresponds to the ``Knowns'' (green quadrant), evaluating the ability to synthesize visual information with parametric knowledge. Crucially, BeyondVisQA targets the ``Known Unknowns'' (gray quadrant), featuring questions that require external information unavailable in the image. This subset rigorously tests the model's self-awareness by examining its ability to refuse to answer when information is insufficient.

Building on the assessment of visual existence, recent works have identified specific failures in the model's awareness of object absence. To probe this, \citet{he2025evaluating} introduced a challenging benchmark (ROHE) focused on \textit{object removal}. They observed that standard evaluations often query irrelevant objects, failing to expose deep-seated biases. By querying images with digitally removed objects, the study revealed that LVLMs frequently claim to ``see'' items that are no longer present. These severe hallucinations arise because the models lack training guidance on visual absence. To address this, they proposed oDPO, a direct preference optimization objective that explicitly guides LVLMs to distinguish between object existence and absence, effectively refining the model's visual self-awareness.

Furthermore, self-awareness must remain robust across diverse domains and noisy scenarios. Addressing the limitations of prior coarse-grained evaluations, \citet{yan2025shale} proposed SHALE, a scalable benchmark designed to assess both faithfulness and factuality hallucinations. Unlike previous works, SHALE employs a hierarchical hallucination induction framework with input perturbations to simulate realistic noisy scenarios. It categorizes hallucinations into fine-grained schemes spanning 12 visual perception aspects and 6 knowledge domains. Their extensive experiments revealed that current LVLMs exhibit high sensitivity to semantic perturbations, indicating a fragility in their self-awareness when facing fine-grained factual and faithful challenges.

\subsubsection{Automated Evaluation Pipelines} 
Automated pipelines facilitate the scaling of hallucination evaluation beyond the constraints of manual curation. To this end, \citet{51} proposed VHTest for automating the generation of Visual Hallucination (VH) instances. This framework utilizes CLIP to retrieve seed instances known as ``CLIP blind pairs,'' defined as images that exhibit high embedding similarity despite being visually distinct. Based on these seeds, the system employs a description model to extract hallucination patterns, which subsequently guide text-to-image models (e.g., DALL-E 3 \cite{betker2023improving}) to synthesize challenging visual samples. However, despite automating image synthesis, VHTest primarily relies on manual construction for QA pairs.

To enable scalable and objective evaluation, \citet{liu2025phd} introduced the PhD benchmark. This framework structures the assessment along two primary axes: visual recognition tasks and interrogation modes. Beyond standard QA scenarios, the authors designed specific adversarial modes, including inquiries with specious or incorrect contexts, and evaluations using counter-common-sense images. By leveraging a ChatGPT-assisted pipeline, this method automatically selects hallucinatory items and generates context-embedded questions, thereby revealing model vulnerability across diverse hallucinatory triggers.

Simultaneously, \citet{yan2025fiha} proposed FIHA to address the need for fine-grained dependency modeling. Unlike methods that rely on costly manual annotations, FIHA autonomously generates Q\&A pairs from images and captions. A key innovation of this framework involves integrating Davidson Scene Graphs (DSG) to structure the evaluation. This design explicitly models the dependencies between distinct visual aspects (e.g., objects, attributes, and relationships) facilitating a comprehensive assessment of how specific hallucinations propagate within the model.



\subsection{Generative Benchmarks}
Generative benchmarks represent a critical paradigm for evaluating Large Vision-Language Models (LVLMs), focusing on the fidelity and factual consistency of their free-form outputs. Unlike discriminative benchmarks that rely on closed-set queries (e.g., binary classification or multiple-choice), generative evaluations probe the model's ability to ground open-ended descriptions in visual reality. This approach is essential for detecting "hallucinations in the wild," where models naturally fabricate non-existent details during captioning or storytelling. However, evaluating generative performance is inherently challenging due to the unstructured nature of natural language and the lack of unique ground truth. To address this, current research has developed diverse methodologies to quantify generative hallucinations. In this section, we categorize these benchmarks into four primary dimensions: (1) \textbf{Hallucination Evaluation}, which utilizes metric-based approaches to calculate the density of fabricated entities in generated text; (2) \textbf{Adversarial Inputs}, which assess model robustness against deceptive visual prompts or misleading instructions; (3) \textbf{Sequential Inputs}, which extend evaluation to the temporal dynamics of video and continuous image sequences; and (4) \textbf{Bidirectional Evaluation}, which verifies the logical self-consistency of LVLMs through cyclic generation (e.g., Image-to-Text-to-Image) and cross-modal alignment.

\subsubsection{Hallucination Evaluation} Caption hallucination assessment with image relevance (CHAIR) \cite{54} is one of the earliest generative methods proposed for evaluating hallucinations in LVLMs. CHAIR includes two variants: ${CHAIR}_i$ and ${CHAIR}_s$. They are defined as follows:
﻿\begin{equation}
\begin{aligned}\text{CHAIR}_i&=\frac{|\{\text{hallucinated objects}\}|}{|\{\text{all objects mentioned}\}|}\\\\\text{CHAIR}_s&=\frac{|\{\text{sentences with hallucinated object}\}|}{|\{\text{ all sentences}\}|}\end{aligned}
\end{equation}
${CHAIR}_i$ calculates the proportion of hallucinated objects to all mentioned objects, and ${CHAIR}_s$ measures the percentage of sentences containing hallucinated objects out of all sentences. To ease the calculation, CHAIR maps words to 80 MSCOCO objects based on a list of synonyms \cite{109}. 

Wang et al. \cite{31} proposed AMBER benchmark to assess the performance of LVLMs for generating hallucinations. In AMBER, each image is annotated with four types of annotations: existence, attribute, relation, and hallucinatory target objects. Existence, attribute and relation refer to the objects existing in the image and their attributes and inter-object relationships. The hallucinatory target objects refer to the hallucinatory objects which may appear in the response of LVLM based on this image. Then, the AMBER designes prompt templates to guide LVLM for answering the questions. Specifically, the counterfactual prompt "Is there a \{hal object\} in this image?" is used to ask whether the hallucinatory target object exists in the image or not. For judgmental questions, AMBER uses accuracy, precision, recall and F1 scores to evaluate hallucinations in response. For generativity questions, AMBER utilizes tool to extract the nouns from the response, and then filters out unnecessary objects to get the list of main objects $R_{o}$. AMBER uses four metrics: $\mathbf{Cover(R)}$, $\mathbf{CHAIR(R)}$, $\mathbf{Hal(R)}$ and $\mathbf{Cog(R)}$ to evaluate generative questions. They can be defined as follows:
﻿\begin{equation}
\mathbf{Cover(R)}=\frac{len(R_{o}\cap A_{o})}{len(A_{o})},
\end{equation}
where $A_o$ denotes the list of ground-truth objects. $\mathbf{Cover(R)}$ measures the completeness of the description of the image by LVLMs.

﻿\begin{equation}
\mathbf{CHAIR(R)}=1-\frac{len(R_{o}\cap A_{o})}{len(A_{o})},
\end{equation}
$\mathbf{CHAIR}$ quantifies the proportion of hallucinated objects in the captions generated by LVLMs.

﻿\begin{equation}
\operatorname{\mathbf{Hal}}(\mathbf{R})=\left\{\begin{array}{ll}
1 & \text { if } \operatorname{CHAIR}(\mathbf{R}) \neq 0 \\
0 & \text { if } \operatorname{CHAIR}(\mathbf{R})=0
\end{array}\right. ,
\end{equation}
$\mathbf{Hal(R)}$ measures the percentage of responses with hallucinations.

﻿\begin{equation}
\mathbf{Cog}(\mathbf{R})=\frac{len(R_{o}\cap H_{o})}{len(R_{o})}
\end{equation}
where $H_o$ denotes the list of hallucinatory target objects, $\mathbf{Cog(R)}$ measures the similarity between hallucinations of LVLM and those conceived by humans.

\subsubsection{Adversarial Inputs} 
When LVLMs receive fraudulent information, they may be misled into generating hallucinations. Qian et al. \cite{44} proposed the MAD-Bench to evaluate model robustness against such inputs. MAD-Bench utilizes GPT-4 to construct six types of adversarial questions based on the COCO dataset \cite{45}, covering aspects like non-existent objects, spatial relationships, and visual confusion. Similarly, CorrelationQA \cite{48} assesses robustness against fraudulent visual inputs. It first generates thirteen meta-category QA pairs with false answers using GPT-4. These answers are then integrated into prompts to generate corresponding fraudulent image instances via Stable Diffusion \cite{49} or OCR techniques, creating a rigorous testbed for visual robustness.


\subsubsection{Sequential Inputs}
Continuous image sequences and videos depict dynamic events, yet benchmarks evaluating LVLM performance in this modality remain scarce. Wang et al. \cite{32} proposed Mementos to evaluate hallucinations in image sequences. This method utilizes GPT-4V to generate detailed event descriptions for each sequence, followed by manual validation to ensure quality. During evaluation, the LVLM is prompted to describe the event in detail. Subsequently, keywords for objects and behaviors are extracted from the response using GPT-4. Mementos then constructs a synonym graph to match these keywords against the ground truth, utilizing recall, precision, and F1 scores to quantify the severity of hallucinations.

Beyond image sequences, evaluating Video-LLMs requires addressing the distinct nature of video captioning. \citet{rawal2025argus} argued that models hallucinate more aggressively in freeform text generation than in multiple-choice tasks. To address this, they introduced ARGUS, a benchmark designed to measure freeform video captioning performance. ARGUS quantifies performance using two metrics. One is the hallucination rate, which tracks incorrect statements regarding video content or temporal relationships. Another is the omission rate, which measures the frequency of missing important descriptive details. Together, these metrics provide a comprehensive view of generation quality.

Focusing on the granularity of temporal hallucinations, \citet{li2025vidhalluc} proposed VidHalluc, a large-scale benchmark comprising 5,002 video pairs. This work highlights that visual encoders often fail to distinguish visually similar yet semantically distinct video pairs. VidHalluc assesses hallucinations across three critical dimensions: action, temporal sequence, and scene transition. Extensive testing on this benchmark reveals that most MLLMs are highly vulnerable to these specific temporal hallucinations. Furthermore, they proposed DINO-HEAL, a training-free mitigation method that incorporates spatial saliency to reweight visual features, effectively reducing hallucination rates.


\subsubsection{Bidirectional Evaluation}
Bidirectional evaluation frameworks assess the robustness of LVLMs by verifying consistency across different modalities (e.g., image-to-text vs. text-to-image), tasks, or against human references.

\paragraph{Cyclic Consistency.}
Drawing inspiration from the game \textit{DrawCeption}, Cao et al. \cite{37} proposed GenCeption, a framework that evaluates hallucinations by measuring semantic drift within an iterative generation cycle. The process begins by prompting the LVLM to generate a detailed description of an original image. Subsequently, a text-to-image model, such as DALL-E \cite{38}, synthesizes a new image based on this description. This process iterates $T$ times to produce a sequence of images. GenCeption quantifies hallucination via the semantic drift metric (GC@$T$), which is defined as:
\begin{equation}
\text{GC@}T := \frac{\sum_{t=1}^T (t \cdot s^{(t)})}{\sum_{t=1}^T t}
\end{equation}
where $s^{(t)}$ denotes the cosine similarity between the $t$-th image and the $(t-1)$-th image. A higher GC@$T$ value indicates superior semantic consistency between the visual and textual modalities, implying that the LVLM introduces fewer hallucinations during the iterative process.

\paragraph{Task-Level Bidirectionality.}
While most benchmarks focus solely on image-to-text generation, Chen et al. \cite{50} introduced the UniHD framework to extend hallucination evaluation to both image-to-text and text-to-image tasks. UniHD utilizes advanced LLMs (e.g., GPT-4V or Gemini) to generate claims for responses (image-to-text) and queries for generation (text-to-image). To ensure factual consistency, UniHD deploys a suite of auxiliary tools to verify these claims, which include object detection, attribute recognition, and scene text recognition. By integrating these tools with an LLM evaluator, UniHD can effectively detect both object-level and factually contradictory hallucinations across bidirectional generation tasks.

\paragraph{Human-Model Alignment.}
Beyond modality cycles, recent work has explored the bidirectional alignment between model generations and human cognition. See Both Ways \cite{katarisee} introduces a human-machine bidirectional evaluation framework to analyze the divergence between LVLM captions and spontaneous human speech. The study compares scores from Human-as-Reference'' and Model-as-Reference,'' revealing significant discrepancies. Humans tend to be selective, prioritizing salient aspects. In contrast, models aim for comprehensive summaries. Furthermore, the study finds that specific categories, such as "nature" and "educational" images, evoke greater human imagination, leading to lower alignment scores. This highlights the necessity of evaluating whether LVLMs can bidirectionally align with human-like selectivity rather than just factual correctness.

\paragraph{Perception-Reasoning Consistency.}
Finally, bidirectional evaluation also entails checking the consistency between visual evidence (Perception) and textual reasoning (Reasoning). Addressing the lack of benchmarks for high-level reasoning hallucinations, PRE-HAL \cite{huang2025visual} constructs a dataset spanning instances, scenes, and relations to systematically evaluate the gap between perception and reasoning. To quantify this gap, the authors propose a detection method based on Dempster-Shafer Theory (DST), which estimates the \textit{evidential conflict} between visual inputs and textual outputs. Experimental results on models like LLaVA-v1.5 and mPLUG-Owl demonstrate that analyzing feature-level conflict offers a robust metric for detecting hallucinations, particularly in complex relation reasoning tasks where visual perception and logical deduction often diverge.

﻿


\section{Future Directions}

\subsection{Unified Theory for Fusion and Uncertainty}
Current hallucination mitigation strategies often rely on empirical heuristics. These methods typically lack foundational theoretical guarantees. Therefore, researchers must establish a unified mathematical framework for multimodal fusion and uncertainty quantification. Hallucinations frequently originate from the failure to weigh conflicting information sources effectively. Models struggle to balance visual evidence against parametric knowledge. Future research should simulate the epistemic uncertainty inherent in each modality. We can formalize the fusion process as a probabilistic inference problem. This approach enables LVLMs to quantify their confidence levels autonomously. For example, a unified theory could theoretically bound the probability of hallucination based on the divergence between visual and textual latent distributions. Furthermore, conformal prediction or Bayesian approximation can allow models to abstain from answering when uncertainty exceeds a safety threshold. We must move from symptom-based fixes to a rigorous theoretical understanding of modality alignment to build trustworthy general-purpose LVLMs.

\subsection{Data-Centric Multimodal Foundation Models Training}
Scaling model parameters yields significant gains, but pre-training data quality remains a primary bottleneck for hallucination. The Multimodal Foundation Models field requires a paradigm shift towards Data-Centric AI. Existing datasets often contain sparse image-text pairs. These pairs frequently describe non-visual concepts or omit visual details, which trains models to hallucinate. Future efforts must prioritize the construction of hallucination-aware datasets. This process involves two key aspects: granularity and negativity. First, training data needs to evolve from coarse-grained image-level captions to fine-grained annotations. This shift grounds textual tokens in specific visual features. Second, we must explicitly incorporate negative samples. These samples describe objects or relationships absent in the image. Teaching the model what is not present is as important as teaching what is present. In addition, the automated process should be able to detect and filter out samples from large amounts of web-crawled data that might cause hallucinations. This strategy fundamentally suppresses hallucinations at the root of the learning process.

\subsection{Hallucination Evaluation Framework for LVLMs} The training data required for LVLMs is massive, for example ViT needing 1.3 million images. Restricted by labor and time costs, researchers usually obtain image-text pairs as training data from the web. Currently, the majority of evaluation benchmarks are open-sourced. If these benchmarks are being used as training data, they lose their role. By using prompt engineering and generative models to produce evaluation benchmarks. For example, leveraging text-to-image generation models like DALL-E 3 to create images, and designing prompts to guide LLMs to generate QA Pairs related to the image content. 

\subsection{Dynamically Evolving Hallucination Correction Framework}At present, most hallucination correction methods rely on an additional training phase for LVLMs. This kind of static correction strategy limits the adaptability and flexibility of model to emerge data types, formats and their underlying contexts. To overcome these limitations, it is particularly important to develop dynamic hallucination correction framework. It not only guarantees that LVLMs continue to learn and adapt from new data and contexts emerge, but also facilitates the ability of model to continuously improve its accuracy, reliability, and generalization in learning process. In addition, it can be realized by integrating contentual learning, incremental learning, meta-learning, feedback loops and other strategies with the hallucination correction methods in the future.

\section{Conclusion}
In this survey, we comprehensively analyze hallucinations in LVLMs and provide insights into the correction methods, assessment benchmarks and future directions. LVLM has the ability to perform advanced functions such as visual question answering, image captioning, cross-modal retrieval and so on. They provide users with a richer interactive experience. However, the hallucination of LVLM reduces the trust of users for the model in practice. To ensure the validity and credibility of LVLM in various applications, it is necessary to improve the reliability and accuracy of LVLM. Therefore, this survey analyzes current hallucination correction strategies based on the causes of hallucination. On the other hand, this survey summarizes the hallucination evaluation benchmarks and divides them into judgmental and generative benchmarks. At the end, we provide three insights into the future direction of hallucination correction, hoping to inspire researchers to address the current shortcomings. Hallucination correction strategies can greatly enhance the application reliability and user trust of LVLMs in various key areas and promote the practical application of AI technology.

\bibliographystyle{ACM-Reference-Format}
\bibliography{mybibliography}




\end{document}